\documentclass[preprint,11pt]{elsarticle}
\usepackage{}
\usepackage{amsmath}
\usepackage{txfonts}
\usepackage{amssymb}
\usepackage{graphicx}
\usepackage{booktabs}
\usepackage{tabularx}
\usepackage{subfigure}
\usepackage{multirow}
\usepackage{url}
\usepackage[figuresright]{rotating}
\usepackage{lipsum}
\usepackage{tikz}
\usepackage{fancybox}
\newtheorem{definition}{\bf Definition}
\newcommand{\PreserveBackslash}[1]{\let\temp=\\#1\let\\=\temp}
\newcolumntype{C}[1]{>{\PreserveBackslash\centering}p{#1}}
\newcolumntype{R}[1]{>{\PreserveBackslash\raggedleft}p{#1}}
\newcolumntype{L}[1]{>{\PreserveBackslash\raggedright}p{#1}}

\journal{Computers \& Industrial Engineering}

\begin{document}

\begin{frontmatter}
\title{A bio-inspired algorithm for fuzzy user equilibrium problem by aid of \textit{Physarum Polycephalum}}
\author[address1]{Yang Liu}
\author[address1]{Xiaoge Zhang}
\author[address1,address2]{Yong Deng\corref{label1}}
\cortext[label1]{Corresponding author: Yong Deng, School of Computer and Information Science, Southwest University, Chongqing, 400715, China. Email address: prof.deng@hotmail.com; yongdeng@nwpu.edu.cn}
\address[address1]{School of Computer and Information Science, Southwest University, Chongqing 400715, China}
\address[address2]{School of Automation, Northwestern Polytechnical University, Xi’an, Shaanxi, 710072, China}

\begin{abstract}
The user equilibrium in traffic assignment problem is based on the fact that travelers choose the minimum-cost path between every origin-destination pair and on the assumption that such a behavior will lead to an equilibrium of the traffic network. In this paper, we consider this problem when the traffic network links are fuzzy cost. Therefore, a \emph{Physarum}-type algorithm is developed to unify the \emph{Physarum} network and the traffic network for taking full of advantage of \textit{Physarum Polycephalum}'s adaptivity in network design to solve the user equilibrium problem. Eventually, some experiments are used to test the performance of this method. The results demonstrate that our approach is competitive when compared with other existing algorithms.

\end{abstract}

\begin{keyword}
Fuzzy user equilibrium; Traffic assignment; \textit{Physarum Polycephalum}; Adaptivity.
\end{keyword}
\end{frontmatter}

\section{Introduction}

The traffic assignment or network equilibrium problem is to predict the steady-state flow of a transportation network \cite{friesz1985transportation,wang1999user,ghatee2009traffic}. Based on this equilibrium, a traffic network can be designed or managed more effectively \cite{farahani2013review,george2013spatio}. There are many literatures for the traffic assignment problem such as multiple route assignment \cite{burrell1968multiple}, probabilistic multipath traffic assignment \cite{dial1971probabilistic} and paired alternative segments for traffic assignment \cite{bar2010traffic}. However, in reality, the state of a transportation network is always determined by independent travelers who only seek to choose the minimum-cost path between every origin-destination pair. According to this phenomenon, in 1952, Wardrop et al. \cite{wardrop1952road} proposed the user equilibrium principle for the traffic network equilibrium problem. Then, a mathematical model of the user equilibrium is developed by Beckmann et al. \cite{beckmann1956studies} in 1956. Based on this model, the dynamic user equilibrium and the stochastic user equilibrium are presented and considered by a lot of researchers \cite{friesz2011approximate,chen2014computation,zhou2012c}.

However, in real world, travelers always can not obtain global information of the traffic network, in which many uncertain events may occur at any moment such as traffic accident. Usually these uncertain events can be taken by fuzzy viewpoints \cite{ghatee2009traffic}. Akiyama et al. \cite{akiyama1993travel} develop a model for route choice behavior due to the fuzzy reasoning approach. Henn \cite{henn2000fuzzy} proposes a fuzzy route choice model by representing travelers with various indices such as risk-taking travelers or risking-averting travelers. Consider the spatial knowledge of individual travelers, Ridwan \cite{ridwan2004fuzzy} suggest a fuzzy preference for travel decisions because some travelers do not follow maximizing principles in route choice. Ghatee et al. \cite{ghatee2009traffic} propose a method based on quasi-logit formulas to obtain a fuzzy equilibrium flow assuming a fuzzy level of travel demand. Many researchers have focused on the fuzzy traffic assignment problem, but, there is not a method dominating the others. Therefore, it is meaningful for us to investigate new method to focus on the fuzzy network equilibrium problem.

Recently, a large, single-celled amoeba-like organism, \emph{Physarum polycephalum}, was found to be adaptively capable of solving many graph theoretical problems such as the shortest path found through a maze \cite{nakagaki2000intelligence,nakagaki2001path,adamatzky2012slime}, path selection in networks \cite{nakagaki2007minimum,zhang2013biologically,zhang2013route}, network design \cite{tero2010rules,adamatzky2012slime11}. In this paper, for taking full of advantage of \textit{Physarum Polycephalum}'s adaptivity in network design, we modify the \emph{Physarum}-type algorithm to unity the \emph{Physarum} network and the traffic network so that they can propagate mutually. By this way, the fuzzy user equilibrium can be approached by \textit{Physarum Polycephalum}. Comparing with other existing algorithms, the main advantage of this algorithm is its adaptivity. To test the performance of this method, some experiments are developed and the results demonstrate that our approach is efficient.

The rest of the paper is organized as follows: in section 2, some preliminaries are presented. In section 3, the proposed method is described. In section 4, experimental results are evaluated. In the final, a brief conclusion is given.

\section{Preliminaries}

Some basic theories are shown in this section, including \emph{Physarum}-type algorithm for shortest path selection, user equilibrium in the traffic network and basic concepts of fuzzy set.

\subsection{\emph{Physarum}-type algorithm for shortest path selection}

The shortest path-selection process of \emph{Physarum polycephalum} is based on the morphogenesis of the tubular structure \cite{nakagaki2001path,tero2007mathematical}: on the one hand, high rate of protoplasmic flow stimulates an increase in tubes diameter, whereas tubes tend to decline at low flow rate. Tube thickness therefore adapts to the flow rate. On the other hand, the decrease of tube thickness is accelerated in the illuminated part of the organism. Thus, the tube structure evolves according to a balance of these mutually antagonistic processes. Based on the observed phenomena of the tube structure’s evolution, a simple \emph{Physarum polycephalum} model, which takes a mathematically simplified and tractable form, is proposed by Tero et al. \cite{tero2007mathematical}.

Using the graphic illustrated in \cite{tero2007mathematical}, the model can be described as follows. Each segment in the diagram represents a section of tube. Two special nodes, which are also called food source nodes, are named $N_1$ and $N_2$, and the other nodes are denoted as $N_3$, $N_4$, $N_5$, and so on. The section of tube between $N_i$ and $N_j$ is denoted as $M_{ij}$. If several tubes connect the same pair of nodes, intermediate nodes will be placed in the center of the tubes to guarantee the uniqueness of the connecting segments. The variable $Q_{ij}$ is used to express the flux through tube $M_{ij}$ from $N_i$ to $N_j$. Assuming the flow along the tube as an approximately Poiseuille flow, the flux $Q_{ij}$ can be expressed as:
\begin{equation}\label{vectmd2}
Q_{ij}=\frac{\pi r_{ij}^4(p_i-p_j)}{8\xi L_{ij}}=\frac{D_{ij}}{L_{ij}}(p_i-p_j)
\end{equation}
where $\xi$ is the viscosity coefficient of the sol. $D_{ij}=\pi r_{ij}^4/8\xi$ is a measure of the tube conductivity. $p_i$ is the pressure at the node $N_i$. $L_{ij}$ is the length of the edge $M_{ij}$.

Assume zero capacity at each node, $\sum_i{Q_{ij}}=0 (j \neq 1,2)$ can be obtained according to the conservation law of sol. For the source node $N_1$ and the sink node $N_2$, $\sum_i{Q_{i1}}=-I_0$ and $\sum_i{Q_{i2}}=I_0$, respectively. $I_0$ is the flux flowing from the source node (or into the sink node). Then the network Poisson equation for the pressure can be obtained as follows:
\begin{equation}\label{vectmd3}
{\sum_i{\frac{D_{ij}}{L_{ij}}(p_i-p_j)}} =\left\{
\begin{array}{l}
\overset{.}{-1}\ \  {for\ j=1,}\\
{+1}\ \ {for\ j=2,}\\
0 \ \ \  {\ otherwise}\\
\end{array}
\right.
\end{equation}
By setting $p_2=0$ as the basic pressure level, all $p_i$'s can be determined by solving above equation system, and each $Q_{ij}=D_{ij}(p_i-p_j)/L_{ij}$ is also obtained.

Experimental observation shows that tubes with larger fluxes are reinforced, while those with smaller fluxes degenerate. To accommodate the adaptive behavior of the tubes, all corresponding conductivities $D_{ij}$'s change in time according to the following equation:
\begin{equation}\label{vectmd4}
\frac{d}{dt}D_{ij}=f(|Q_{ij}|)-aD_{ij}
\end{equation}
where $a$ is a decay rate of the tube. The functional form $f(|Q_{ij}|)$ is generally given by $f(|Q_{ij}|)=|Q_{ij}|$ for the sake of simplicity \cite{tero2007mathematical}. To solve the adaption of Eq. (\ref{vectmd4}), a semi-implicit scheme is used as follows:
\begin{equation}\label{vectmd5}
\frac{D_{ij}^{n+1}-D_{ij}^n}{\delta t}=|Q_{ij}^n|-D_{ij}^{n+1}
\end{equation}
where $\delta t$ is a time mesh size and the upper index $n$ indicates a time step. Hence, the time-varying conductivity $D_{ij}$ generates the state of the system to make the shortest path emerge.

\subsection{User equilibrium in the traffic network}

\shadowbox{
\begin{minipage}{4.5 in}
\textbf{Nomenclature}\\
\\
$N:$ the set of network nodes\\
$A:$ the set of network links $\{a\}$\\
$\Pi:$ the set of origin-destination ($O-D$) nodes $\{(o,d)\}$\\
$P_{(o,d)}:$ the set of paths from $o$ to $d$ for each $(o,d)\in \Pi$\\
$q_{(o,d)}$: the demand of trips through $(o,d)\in \Pi$\\
$x^a:$ the total flow through link $a\in A$\\
$u^a:$ the capacity of link $a\in A$\\
$c_0^a:$ the free-flow cost of link $a\in A$\\
$\delta_{(o,d)}^{p,a}=\left\{ \begin{array}{l} 1,\ \ \ a\in p,\\ 0,\ \ \ a\notin p.\\ \end{array}\right.$ \ \ $\forall a\in A, \forall p \in P_{(o,d)}$\\
$c^a:$ the cost on link $a$, $c^a=c^a(x^a,c_0^a)$\\
$f_{(o,d)}^p:$ the traffic flow of path $p\in P_{(o,d)}$\\
$c_{(o,d)}^p:$ the cost of path $p\in P_{(o,d)}$, $c_{(o,d)}^p=\sum_{a\in A}c^a\delta_{(o,d)}^{p,a}$\\
\end{minipage} }

In 1952, Wardrop et al. \cite{wardrop1952road} proposed the User equilibrium principle: any travelers can not decrease themselves' cost by changing travel route when the traffic system is equilibrium. According to this principle, a flow-cost formula was developed by Beckmann et al. \cite{beckmann1956studies}:
\begin{equation}\label{vectmd6}
\mu_{(o,d)} -c_{(o,d)}^p= \left\{
\begin{array}{l}
=0,\ \ \ f_{(o,d)}^p>0,\forall p\in P_{(o,d)},\forall (o,d) \in \Pi,\\
\leq0,\ \ \ f_{(o,d)}^p=0,\forall p\in P_{(o,d)},\forall (o,d) \in \Pi.\\
\end{array}
\right.
\end{equation}
where $\mu_{(o,d)}$ is the minimum cost of each $(o,d)$ in the network equilibrium. Assuming that the cost of each link is only associated with the flow of that link and the cost is strictly increasing with the flow increasing, Eq. \ref{vectmd6} can be transformed into the mathematical model shown as follows:

\begin{equation}\label{vectmd7}
\min Z(w)=\sum_{a\in A}\int_0^{x^a}c^a(w)dw,
\end{equation}
\begin{equation}\label{vectmd8}
s.t.\left\{
\begin{array}{l}
\sum_{p\in P_{(o,d)}f_{(o,d)}^p}=q_{(o,d)}\ \ \ \forall (o,d)\in \Pi,\\
x^a=\sum_{(o,d)\in \Pi}\sum_{p\in P_{(o,d)}}f_{(o,d)}^p \delta_{(o,d)}^{p,a}\ \ \ \forall a\in A,\\
f_{(o,d)}^p\geq 0.
\end{array}
\right.
\end{equation}

It can be proved that there exists one unique solution for Eq. \ref{vectmd7} \cite{smith1979existence}. Hence, The user equilibrium problem can be solved by solving a mathematical optimization problem as Eq. \ref{vectmd7} and Eq. \ref{vectmd8}.

\subsection{Fuzzy set}

Fuzzy set proposed by Zadeh \cite{Zadeh1965} in 1965 is widely used in many fields such as statistics \cite{yang2010fuzzy,giri2014fuzzy}, computer programming \cite{azadeh2012flexible,hu2011reversed}, engineering and experimental science \cite{nguyen2011computing,Deng2011a}. Based on this theory, The concept of fuzzy number was first used by Nahmias in the United States and by Dubois and Prade in France in the late 1970s. In this paper, the triangular fuzzy number will be used. Therefore, according to \cite{Kauffman1991, Ezzati2010},some basic definitions of fuzzy set and fuzzy number are given as follows.

\begin{definition}
A fuzzy set $\widetilde{A}$ defined on a universe X may be expressed as:
\begin{eqnarray}
\widetilde{A} = \left\{ {\left\langle {x,\mu _{\widetilde{A}} \left( x \right)}
\right\rangle \left| {x \in X} \right.} \right\}
\end{eqnarray}
where \(\mu _{\widetilde{A}} \to \left[ {0,1} \right] \) is the membership
function of $\widetilde{A}$. The membership value $\mu _{\widetilde{A}} \left( x \right)$
describes the degree of $x \in X$ in $\widetilde{A}$.
\end{definition}

\begin{definition}
A fuzzy set $\widetilde{A}$ of X is normal iff
\(
\sup _{x \in X} \mu _{\widetilde{A}} \left( x \right) = 1
\).
\end{definition}

\begin{definition}
A fuzzy set $\widetilde{A}$ of X is convex iff
\(
\mu _{\widetilde{A}} \left( {\lambda x + \left( {1 - \lambda } \right)y} \right) \ge \left( {\mu _{\widetilde{A}} \left( x \right) \wedge \mu _{\widetilde{A}} \left( y \right)} \right),\;\\ \forall x,y \in X,\forall \lambda  \in \left[ {0,1} \right]
\), where $\wedge$ denotes the minimum operator.
\end{definition}

\begin{definition}
A fuzzy set $\widetilde{A}$ is a fuzzy number iff $\widetilde{A}$ is normal and convex on X.
\end{definition}

\begin{figure}[!ht]
\centering
\includegraphics[width=2.8in]{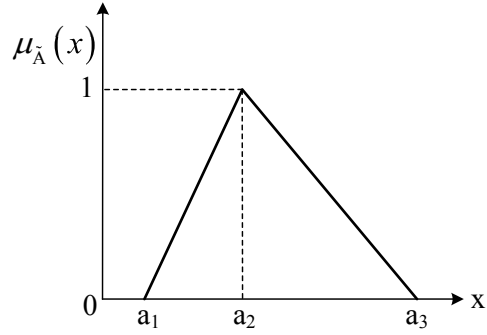}
\caption{A triangular fuzzy number $\widetilde{A}$}
\label{fig_triangular}
\end{figure}

\begin{definition}
A triangular fuzzy number $\widetilde{A}$ is a fuzzy number with a piecewise linear membership function $\mu _{\widetilde{A}}$ defined by:
\begin{eqnarray}
\mu _{\widetilde{A}}  = \left\{ \begin{array}{l}
 0,\quad \quad \quad \; x \le a_1  \\
 \frac{{x - a_1 }}{{a_2  - a_1 }},\quad \;\;a_1  \le x \le a_2  \\
 \frac{{a_3  - x}}{{a_3  - a_2 }},\quad \;\;a_2  \le x \le a_3  \\
 0,\quad \quad \quad \; a_3  \le x \\
 \end{array} \right.
\end{eqnarray}
which can be denoted as a triplet \( \left( a_1, a_2, a_3 \right) \). A triangular fuzzy number $\widetilde{A}$ in the universe set $X$ that conforms to this definition shown in Fig. \ref{fig_triangular}.
\end{definition}

Based on \cite{Giachetti1997,Chen1994}, fuzzy arithmetic on triangular is shown as follows.

\begin{definition}
Assuming that both \(\widetilde{A} = \left( a_1, a_2, a_3 \right) \) and \(\widetilde{B} = \left( b_1, b_2, b_3 \right) \) are triangular numbers, then the basic fuzzy operations are:
\begin{eqnarray}\label{eqn_TriArithmetic}
&& \widetilde{A} \oplus \widetilde{B} = \left( {a_1  + b_1 ,a_2  + b_2 ,a_3  + b_3 }  \right ) \quad {\rm{for \; addition,}} \\
&& \widetilde{A} \ominus \widetilde{B} = \left( {a_1  - b_3 ,a_2  - b_2 ,a_3  - b_1 } \right)\quad {\rm{for \; subtraction,}} \\
&& \widetilde{A} \otimes \widetilde{B} = \left( {a_1  \times b_1 ,a_2  \times b_2 ,a_3  \times b_3 } \right)  \quad {\rm{for \; multipulation,}}\\
&& \widetilde{A} \oslash \widetilde{B}  = \left( {a_1 \; / \; b_3\, ,a_2 \; / \; b_2\, ,a_3  \; / \; b_1 \; } \right) \quad {\rm{for \; division.}}
\end{eqnarray}
\end{definition}

\begin{definition}
Assuming that both \(\widetilde{A} = \left( a_1, a_2, a_3, a_4 \right) \) and \(\widetilde{B} = \left( b_1, b_2, b_3, b_4 \right) \) are trapezoidal numbers, then the basic fuzzy operations are:
\begin{eqnarray}
&& \widetilde{A} \oplus \widetilde{B} = \left( {a_1  + b_1 ,a_2  + b_2 ,a_3  + b_3 ,a_4  + b_4 }  \right ) \quad {\rm{for \; addition,}} \\
&& \widetilde{A} \ominus \widetilde{B} = \left( {a_1  - b_4 ,a_2  - b_3 ,a_3  - b_2, a_4 - b_1 } \right)\quad {\rm{for \; subtraction,}} \\
&& \widetilde{A} \otimes \widetilde{B} = \left( {a_1  \times b_1 ,a_2  \times b_2 ,a_3  \times b_3, a_4 \times b_4 } \right)  \quad {\rm{for \; multipulation,}}\\
&& \widetilde{A} \oslash  \widetilde{B}  = \left( {a_1 \; / \; b_4\, ,a_2 \; / \; b_3\, ,a_3  \; / \; b_2\, ,a_4  \; / \; b_1 \;} \right) \quad {\rm{for \; division.}}
\end{eqnarray}
\end{definition}

For example, let \(\widetilde{A} = \left( 8,10,12 \right) \) and \(\widetilde{B} = \left( 4, 5, 6 \right) \) be two triangular fuzzy numbers. Based on Eqs. \ref{eqn_TriArithmetic}, four basic operations can be derived as:
\begin{eqnarray*}
&& \widetilde{A} \oplus \widetilde{B} = \left( {8,10,12} \right) \oplus \left( {4,5,6} \right) = \left( {12,15,18} \right). \\
&& \widetilde{A} \ominus \widetilde{B} = \left( {8,10,12} \right) \ominus \left( {4,5,6} \right) = \left( {2,5,8} \right).\\
&& \widetilde{A} \otimes \widetilde{B} = \left( {8,10,12} \right) \otimes \left( {4,5,6} \right) = \left( {32,50,72} \right).\\
&& \widetilde{A} \oslash  \widetilde{B}  = \left( {8,10,12} \right) \oslash \left( {4,5,6} \right) = \left( {8/6,2,3} \right).
\end{eqnarray*}

\begin{figure}[!ht]
\centering
\includegraphics[width=5.5in]{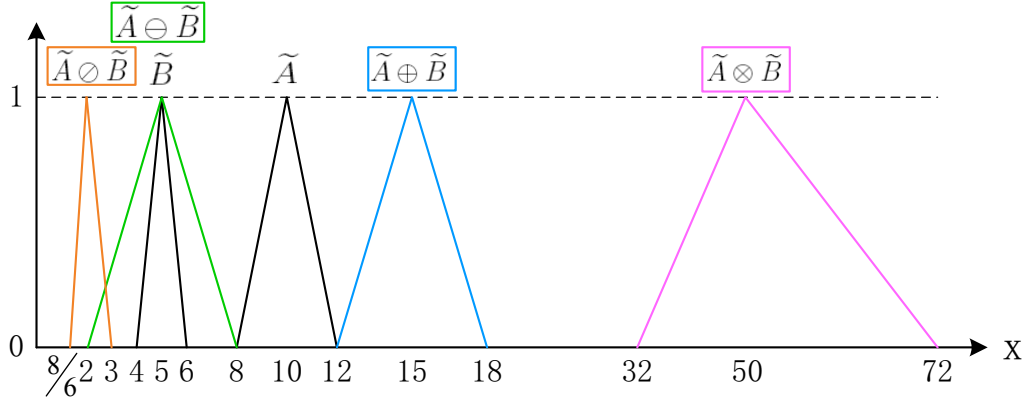}
\caption{An example of fuzzy arithmetic operations on triangular fuzzy numbers \(\widetilde{A} = \left( 8,10,12 \right) \) and \(\widetilde{B} = \left( 4, 5, 6 \right) \)}
\label{fig_farithmetic}
\end{figure}

The results of the above operations are depicted in Fig. \ref{fig_farithmetic}.

Recently, fuzzy distance, as a measure of distance between two fuzzy numbers, has gained much attention from researchers and been widely applied in data analysis, classification, and so on \cite{guha2010new,sadi2010application}. In this paper, the $Dis_{p,q}$-distance proposed in \cite{Gildeh2001} is adopted to measure the difference between two fuzzy numbers.

\begin{definition}
The $Dis_{p,q}$-distance, indexed by parameters $1 < p < \infty$ and $0 < q < 1$, between two fuzzy numbers $\widetilde{A}$ and $\widetilde{B}$ is a nonnegative function given by \cite{Gildeh2001, Mahdavi2009}:
\begin{eqnarray}
Dis_{p,q} \left( {\widetilde{A}{\rm{,}}\widetilde{B}} \right) = \left\{ \begin{array}{l}
 \left[ {\left( {1 - q} \right)\int_0^1 {\left| {A_\alpha ^ -   - B_\alpha ^ -  } \right|^p {\mathop{\rm d}\nolimits} \alpha }  + q\int_0^1 {\left| {A_\alpha ^ +   - B_\alpha ^ +  } \right|^p {\mathop{\rm d}\nolimits} \alpha } } \right]^{{1 \mathord{\left/
 {\vphantom {1 p}} \right.
 \kern-\nulldelimiterspace} p}} ,\quad p < \infty , \\
 \left( {1 - q} \right)\mathop {\sup }\limits_{0 < \alpha  \le 1} \left( {\left| {A_\alpha ^ -   - B_\alpha ^ -  } \right|} \right) + q\mathop {\inf }\limits_{0 < \alpha  \le 1} \left( {\left| {A_\alpha ^ +   - B_\alpha ^ +  } \right|} \right),\quad \quad \;\;p = \infty . \\
 \end{array} \right.
\end{eqnarray}
\end{definition}

The analytical properties of $Dis_{p,q}$ depend on the first parameter $p$, while the second parameter $q$ of $Dis_{p,q}$ characterizes the subjective weight attributed to the end points of the support. Having $q$ close to 1 results in considering the right side of the support of the fuzzy numbers more favorably. Since the significance of the end points of the support of the fuzzy numbers is assumed to be same, the $q = (1/2)$ is adopted in this paper.

According to studies by Mahdavi et al. \cite{Mahdavi2009,Hassanzadeh2011}, with $p=2$ and $q=(1/2)$, the general form of fuzzy distance $Dis_{p,q}$ can be converted into different forms, as two fuzzy numbers $\widetilde{A}$ and $\widetilde{B}$ take different types.

For triangular fuzzy numbers \(\widetilde{A} = \left( a_1, a_2, a_3 \right) \) and \(\widetilde{B} = \left( b_1, b_2, b_3 \right) \), the fuzzy distance between them can be represented as:

\begin{eqnarray}
Dis \left( {\widetilde{A}{\rm{,}}\widetilde{B}} \right) = \sqrt {\frac{1}{6}\left[ {\sum\limits_{i = 1}^3 {\left( {b_i  - a_i } \right)^2 }  + \left( {b_2  - a_2 } \right)^2  + \sum\limits_{i \in \left\{ {1,2} \right\}} {\left( {b_i  - a_i } \right)\left( {b_{i + 1}  - a_{i + 1} } \right)} } \right]}
\end{eqnarray}

\section{Proposed method}


\subsection{Fuzzy user equilibrium}

In the model of traditional user equilibrium, a basic assumption is that all travelers know the global information of the traffic network. Based on this, every one can make the decision having no conflict with each other. In reality, however, because of some uncertain events or local information, travelers always need to make the decision according to the fuzzy information of the traffic network. In general, the fuzzy information can be divided into three types \cite{ghatee2009traffic}: inexact travel cost, unsure network topology and imprecise travel demand. In this paper, we consider the user equilibrium problem in traffic assignment with fuzzy travel cost or fuzzy user equilibrium problem. According to section 2.2, the fuzzy user equilibrium can be stated as follows:

\begin{equation}\label{vectmd9}
\min \widetilde{Z}(w)=\sum_{a\in A}\int_0^{x^a}\widetilde{c}^a(w)dw,
\end{equation}
\begin{equation}\label{vectmd10}
s.t.\left\{
\begin{array}{l}
\sum_{p\in P_{(o,d)}f_{(o,d)}^p}=q_{(o,d)}\ \ \ \forall (o,d)\in \Pi,\\
x^a=\sum_{(o,d)\in \Pi}\sum_{p\in P_{(o,d)}}f_{(o,d)}^p \delta_{(o,d)}^{p,a}\ \ \ \forall a\in A,\\
f_{(o,d)}^p\geq 0.
\end{array}
\right.
\end{equation}
where $\widetilde{c}^a(\cdot)$ is the fuzzy cost function. It denotes the inaccuracies of perceived time of travelers. For example, one traveler may have a larger perceived travel time of one path than its real travel time while some travelers increase their speed through this paths for some special reasons and a lower if a car accident occurs on this path. For the sake of simplicity, we assign $\widetilde{c}^a$ with a triangular fuzzy number. To associate $\widetilde{c}^a$ with $x^a$, we use $\alpha_l$ and $\alpha_r$ to extend $x^a$. Therefore, $\widetilde{c}^a$ can be calculated as follows:
\begin{equation}\label{vectmd11}
\widetilde{c}^a=(c^a[(1-\alpha_l)x^a],c^a(x^a),c^a[(1+\alpha_r)x^a])
\end{equation}
where $c^a[(1-\alpha_l)x^a]$ is the left limit and $c^a[(1+\alpha_r)x^a]$ is the right limit of $c^a$, respectively. The advantage of using such a strategy is it considers link capacity as an effective factor in traffic assignment \cite{ramazani2011fuzzy}.

\subsection{\emph{Physarum}-type algorithm for fuzzy user equilibrium algorithm}

According to the network structure, \textit{Physarum Polycephalum} is able to make full use of its protoplasm (flow) for building a new network (\emph{Physarum} network) based on its adaptivity \cite{tero2010rules}. To take full of advantage of this feature for fuzzy user equilibrium problem, it is necessary for us to find out the similarities and differences between the \emph{Physarum} network and the traffic network, and then to find a way to unify the \emph{Physarum} network and the traffic network. Therefore, some modifications of \emph{Physarum}-type algorithm should be carried out.

There are many similar properties between the \textit{Physarum} network and the traffic network. As a result, we can treat the links in the traffic network as the tubes in \textit{Physarum} network, the traffic flow as the protoplasm, the traffic nodes as the food sources and the cost as the distance. Meanwhile, there are some differences between them. For example, in reality, the traffic nodes have many their own features such as education center, political centers and transportation hubs. Besides, the traffic flow is always determined by many factors such as the choices of travelers, the accidents and the transport facilities. While in the \textit{Physarum} network, the protoplasm in the tubes only flow from the high-pressure node to the low-pressure node.


Based on these similarities and differences, we rewrite some formulas of \emph{Physarum}-type algorithm to unify the \emph{Physarum} network and the traffic network. Firstly, consider the distance of tubes and the cost of links:
\begin{equation}\label{vectmd12}
\widetilde{L}_{ij}=\widetilde{c}^a, \ \ \ i,j\in N
\end{equation}
where arc from node $i$ to node $j$ is equal to link $a$. Like \cite{zhang2013biologically}, fuzzy cost $\widetilde{c}^a$ can be denoted as $(c_l^a,c^a,c_r^a)$. Then, Eq. \ref{vectmd3} can be rewrite as follows:
\begin{equation}\label{vectmd13}
\begin{split}
{\sum_i{\frac{D_{ij}}{c_l^a}(p_i^l-p_j^l)}} =\left\{
\begin{array}{l}
\overset{.}{-q_{(o,d)}}\ \  {for\ j=1,}\\
{+q_{(o,d)}}\ \ {for\ j=2,}\\
0 \ \ \  {\ otherwise.}\\
\end{array}
\right.\\
{\sum_i{\frac{D_{ij}}{c^a}(p_i-p_j)}} =\left\{
\begin{array}{l}
\overset{.}{-q_{(o,d)}}\ \  {for\ j=1,}\\
{+q_{(o,d)}}\ \ {for\ j=2,}\\
0 \ \ \  {\ otherwise.}\\
\end{array}
\right.\\
{\sum_i{\frac{D_{ij}}{c_r^a}(p_i^r-p_j^r)}} =\left\{
\begin{array}{l}
\overset{.}{-q_{(o,d)}}\ \  {for\ j=1,}\\
{+q_{(o,d)}}\ \ {for\ j=2,}\\
0 \ \ \  {\ otherwise.}\\
\end{array}
\right.\\
\end{split}
\end{equation}
where $\widetilde{p}_i=(p_i^l,p_i,p_i^r),i \in N$ is associated with the fuzzy cost $\widetilde{c}^a$. According to $\widetilde{p}_i$, the $\widetilde{p}_{(o,d)}$ can be obtained for each $(o,d)$ pair. Then, the globe pressure $\widetilde{p}_{(O,D)}$ can be calculated as follows:
\begin{equation}\label{vectmd17}
\widetilde{p}_{(O,D)}=\sum_{(o,d)\in \Pi}\widetilde{p}_{(o,d)}
\end{equation}
Next, $x^a$ can be calculated as follows:
\begin{equation}\label{vectmd14}
x^a=D_{ij}\times Dis(\frac{\widetilde{p}_i}{\widetilde{c}^a},\frac{\widetilde{p}_j}{\widetilde{c}^a}),\ \ \ \widetilde{p}_i, \widetilde{p}_j \in \widetilde{p}_{(O,D)}
\end{equation}
where $Dis(\cdot)$ is the function of fuzzy distance measure. Finally, according to Eq. \ref{vectmd5} and $x^a$, $D_{ij}$ is obtain as follows:
\begin{equation}\label{vectmd15}
\frac{D_{ij}^{n+1}-D_{ij}^n}{\delta t}=|(x^a)^n|-D_{ij}^{n+1}
\end{equation}
where $\delta t$ is a time mesh size and the upper index $n$ indicates a time step.

\section{Experimental results}

In this section, we briefly illustrate the efficiency of the proposed algorithm by studying on some sample networks.
\subsection{A test problem of Ramazani}

\begin{figure}[!ht]
\centering
\includegraphics[width=2.5in]{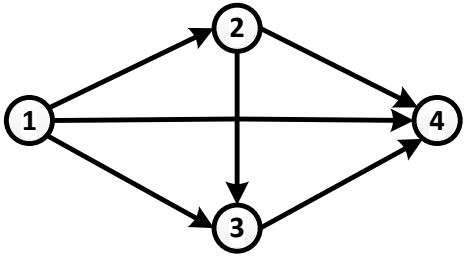}
\caption{A small network with 4 nodes}
\label{example1}
\end{figure}

The first experiment involves a test problem introduced by Ramazani et al. \cite{ramazani2011fuzzy}. It is based on a network with 4 nodes, 6 links and one origin-destination travel demands shown as Fig. \ref{example1}. According to the well known cost function represented by the US Bureau of Public Roads \cite{manual1964us}, the cost of link $a$ can be calculated as follows:

\begin{equation}\label{vectmd16}
c^a(x)=c_0^a\Bigl(1+\alpha (\frac{x^a}{u^a})^\beta \Bigr)
\end{equation}
where $c_0^a$ and $u^a$ are the free-flow cost and the capacity of link $a\in A$ depicted in Table \ref{Information}. Parameters $\alpha$ and $\beta$ are fixed values (usual values are $\alpha=0.15$ and $\beta=4$).

\begin{table}[!h]
\tabcolsep 0pt
\caption{Example network free-flow cost and the capacity of links}
\vspace*{-12pt}
\begin{center}
\def\temptablewidth{0.9\textwidth}
{\rule{\temptablewidth}{1pt}}
\begin{tabular*}{\temptablewidth}{@{\extracolsep{\fill}}lllllll}
            Path        &  (1,2) & (1,3) & (2,3) & (2,4) & (1,4) & (3,4)  \\
    \midrule
    $c_0^a$   &  4    &  5 & 7 & 7 & 17 & 7    \\
    $u^a$   &  200  &  150 & 250 & 250 & 300 & 250     \\
\end{tabular*}
{\rule{\temptablewidth}{1pt}}
\end{center}
\label{Information}
\end{table}

\begin{figure}[!ht]
\centering
\includegraphics[width=4in]{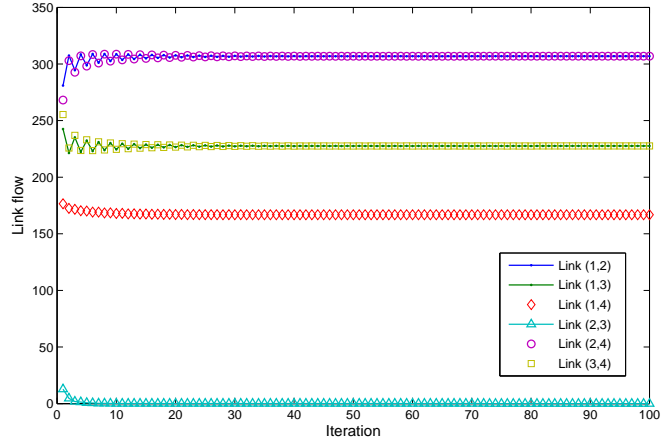}
\caption{Small network: propagation of the travel flows along each link}
\label{100ite}
\end{figure}

\begin{table}[!h]
\tabcolsep 0pt
\caption{Final assignment results of traffic volume for 100 iterations}
\vspace*{-12pt}
\begin{center}
\def\temptablewidth{0.9\textwidth}
{\rule{\temptablewidth}{1pt}}
\begin{tabular*}{\temptablewidth}{@{\extracolsep{\fill}}lllllll}
            Links        &  (1,2) & (1,3) & (2,3) & (2,4) & (1,4) & (3,4)  \\
    \midrule
    FITA \cite{ramazani2011fuzzy}   &  287    &  217 & 0 & 287 & 196 & 217   \\
    PA    &  306  &  227 & 0 & 306 & 167 & 227   \\
\end{tabular*}
{\rule{\temptablewidth}{1pt}}
\end{center}
\label{Re1}
\end{table}

The demand for origin-destination $1-4$ is 700, namely $q_{(1,4)}=700$. Parameter $\alpha_l$ and $\alpha_r$ are assumed to be $\alpha_l=\alpha_r=0.2$. The results of assignment after \emph{Physarum}-type algorithm (PA) are shown as Fig. \ref{100ite} and Table \ref{Re1}. Consider this equilibrium state, the fuzzy cost of each path can be calculated according to Eq. \ref{vectmd11}. For comparing these results, we use the method in \cite{deng2012fuzzy,hassanzadeh2013genetic} to transform the fuzzy cost to crisp number shown as Table \ref{re2}. Therefore, PA is more efficient than FITA.

\begin{table}[!h]
\tabcolsep 0pt
\caption{Example network free-flow cost and the capacity of links}
\vspace*{-12pt}
\begin{center}
\def\temptablewidth{0.9\textwidth}
{\rule{\temptablewidth}{1pt}}
\begin{tabular*}{\temptablewidth}{@{\extracolsep{\fill}}llllll}
            Method & Route        &  $1\to 2\to 4$ & $1\to 4$ & $1\to 3\to 4$  \\
    \midrule
    \multirow{2}{*}{Hassanzadeh's method \cite{hassanzadeh2013genetic}} & FITA   &  50.48    &  55.39 & 51.85  \\
    & PA   &  55.16  &  54.62 & 54.64     \\
    \multirow{2}{*}{Deng's method \cite{deng2012fuzzy}} & FITA   &  15.72    &  17.50 & 16.19  \\
    & PA   &  17.10  &  17.26 & 17.02     \\
\end{tabular*}
{\rule{\temptablewidth}{1pt}}
\end{center}
\label{re2}
\end{table}

\subsection{A test problem of Ghatee}

\begin{figure}[!ht]
\centering
\includegraphics[width=4in]{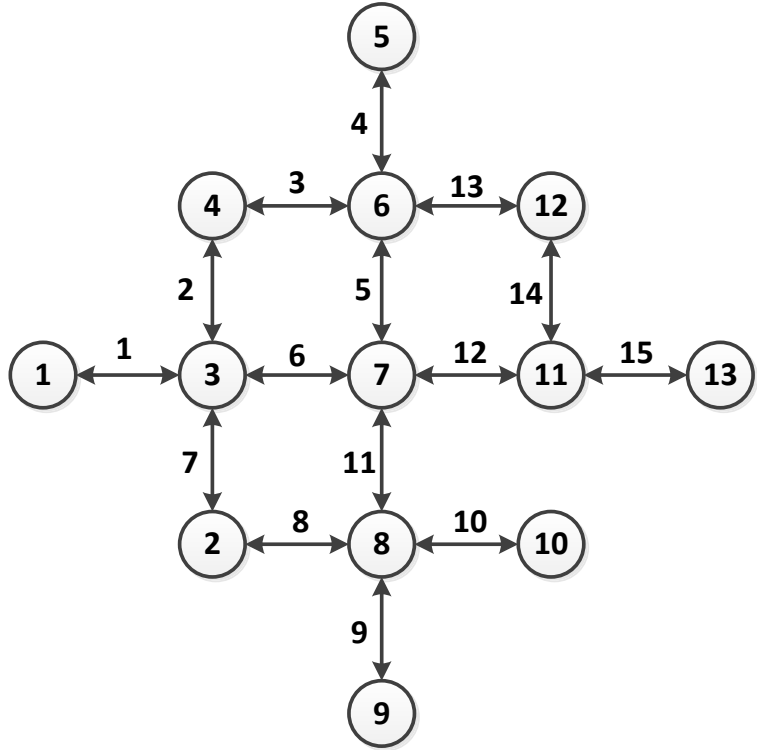}
\caption{A 13 nodes, 15 links and 5 junctions network}
\label{example2}
\end{figure}

\begin{table}[!h]
\tabcolsep 0pt
\caption{The capacity and free-flow cost of network links (depicted in Fig. \ref{example2})}
\vspace*{-12pt}
\begin{center}
\def\temptablewidth{0.9\textwidth}
{\rule{\temptablewidth}{1pt}}
\begin{tabular*}{\temptablewidth}{@{\extracolsep{\fill}}lllll}
            Ind.  &  Node $i$ & Node $j$ & Free-F. C. & Link Cap.  \\
    \midrule
    1  &  1 & 3 & 4 & 252  \\
    2  &  3 & 4 & 13 & 415  \\
    3  &  4 & 6 & 13 & 413  \\
    4  &  5 & 6 & 21 & 175  \\
    5  &  6 & 7 & 8 & 174  \\
    6  &  3 & 7 & 21 & 367  \\
    7  &  2 & 3 & 17 & 423  \\
    8  &  2 & 8 & 20 & 189  \\
    9  &  8 & 9 & 18 & 277  \\
    10  &  8 & 10 & 10 & 351  \\
    11  &  7 & 8 & 8 & 401  \\
    12  &  7 & 11 & 10 & 265  \\
    13  &  6 & 12 & 7 & 90  \\
    14  &  11 & 12 & 11 & 139  \\
    15  &  11 & 13 & 21 & 442  \\
\end{tabular*}
{\rule{\temptablewidth}{1pt}}
\end{center}
\label{exa2}
\end{table}

\begin{table}[!h]
\tabcolsep 0pt
\caption{The equilibrium flow of network links (depicted in Fig. \ref{example2})}
\vspace*{-12pt}
\begin{center}
\def\temptablewidth{0.9\textwidth}
{\rule{\temptablewidth}{1pt}}
\begin{tabular*}{\temptablewidth}{@{\extracolsep{\fill}}llll}
            Ind.  &  Node $i$ & Node $j$ & Flow \\
    \midrule
    1  &  1 & 3 & 400.00  \\
    2  &  3 & 4 & 0.00  \\
    3  &  4 & 6 & 0.00  \\
    4  &  5 & 6 & 450.00  \\
    5  &  6 & 7 & 300.00  \\
    6  &  3 & 7 & 353.67  \\
    7  &  2 & 3 & 46.33  \\
    8  &  2 & 8 & 46.33 \\
    9  &  8 & 9 & 250.00  \\
    10  &  8 & 10 & 350.00  \\
    11  &  7 & 8 & 553.67 \\
    12  &  7 & 11 & 100.00  \\
    13  &  6 & 12 & 150.00  \\
    14  &  11 & 12 & 150.00  \\
    15  &  11 & 13 & 250.00  \\
\end{tabular*}
{\rule{\temptablewidth}{1pt}}
\end{center}
\label{exa2flow}
\end{table}

\begin{table}[!h]
\tabcolsep 0pt
\caption{The results of the fuzzy cost of different $(o,d)$ pair}
\vspace*{-12pt}
\begin{center}
\def\temptablewidth{1.2\textwidth}
{\rule{\temptablewidth}{1pt}}
\begin{tabular*}{\temptablewidth}{@{\extracolsep{\fill}}llllll}
            $(o,d)$ & Paths        &  Fuzzy cost  & Deng's method\\
    \midrule
    \multirow{2}{*}{(1,9)} & $1 \to 3\to 2\to 8\to 9$   &  $(61.2985,64.6115,70.6359)$ & 65.0634  \\
    & $1\to 3\to 7\to 8\to 9$   &  $(56.1930,63.6782,77.2894)$ & 64.6992 \\

    \multirow{2}{*}{(1,10)} & $1 \to 3\to 2\to 8\to 10$   &  $(53.1721,56.3030,61.9963)$ & 56.7301  \\
    & $1\to 3\to 7\to 8\to 10$   &  $(48.0666,55.3697,68.6498)$ & 56.3659 \\

    \multirow{1}{*}{(1,13)} & $1 \to 3\to 7\to 11\to 13$   &  $(58.8174,62.8783,70.2628)$ & 63.4322 \\

    \multirow{1}{*}{(5,9)} & $5 \to 6\to 7\to 8\to 9$   &  $(118.2751,209.4803,375.3303)$ & 221.9211 \\

    \multirow{1}{*}{(5,10)} & $5 \to 6\to 7\to 8\to 10$   &  $(110.1488,201.1718,366.6906)$ & 213.5878  \\

    \multirow{2}{*}{(5,13)} & $5 \to 6\to 12\to 11\to 13$   &  $(120.7787,208.3855,366.6922)$ & 220.3355 \\
    & $5\to 6\to 7\to 11\to 13 $   &  $(120.8995,208.6804,368.3037)$  & 220.6541\\

\end{tabular*}
{\rule{\temptablewidth}{1pt}}
\end{center}
\label{exa2re}
\end{table}

The second experiment involves a test problem introduced by Ramazani et al. \cite{ghatee2009traffic}. It is based on a network with 13 nodes, 15 links and five junctions depicted in Fig. \ref{example2}. The capacity and free-flow cost of network links are presented in Table \ref{exa2}. Assume six $(o,d)$ pairs $\{(1,9),(1,10),(1,13),(5,9),(5,10),(5,13)\}$ with demand $\{100,200,100,150,150,150\}$, the results of the equilibrium flows and the equilibrium path data of each $(o,d)$ pair are presents in Table \ref{exa2flow} and Table \ref{exa2re}, respectively.
Table \ref{exa2flow}

\section{Conclusion}

In this paper, we propose a \emph{Physarum} network mode to address the fuzzy user equilibrium problem. We modify the \emph{Physarum}-type algorithm to build a relationship between the \emph{Physarum} network and the traffic network so that they can propagate mutually. By this way, an equilibrium state occurs approaching the solution of the fuzzy user equilibrium problem. To test the performance of the proposed method, some experiments are developed. The results demonstrate the feasibility and effectiveness of the proposed algorithm.

\section{Acknowledgments}

The author greatly appreciate the reviews' suggestions. The work is partially supported by National Natural Science Foundation of China (Grant No. 61174022), Specialized Research Fund for the Doctoral Program of Higher Education (Grant No. 20131102130002), R\&D Program of China (2012BAH07B01), National High Technology Research and Development Program of China (863 Program) (Grant No. 2013AA013801), the open funding project of State Key Laboratory of Virtual Reality Technology and Systems, Beihang University (Grant No.BUAA-VR-14KF-02).

\bibliographystyle{model1-num-names}
\bibliography{dssbib}
\end{document}